\documentclass[10pt]{article}

\usepackage{amsmath,amsfonts,bm}

\def\eqref#1{equation~\ref{#1}}

\def\1{\bm{1}}

\DeclareMathAlphabet{\mathsfit}{\encodingdefault}{\sfdefault}{m}{sl}
\SetMathAlphabet{\mathsfit}{bold}{\encodingdefault}{\sfdefault}{bx}{n}

\usepackage{hyperref}
\usepackage{url}
\usepackage{todonotes}
\usepackage{nicefrac}
\usepackage{booktabs}
\usepackage{cleveref}
\usepackage[whole]{bxcjkjatype}  %
\usepackage{graphicx}
\usepackage{enumitem}
\usepackage{pifont}
\usepackage{natbib}
\usepackage{authblk}
\usepackage{rotating}
\usepackage{tabularx}
\usepackage{ltablex}

\usepackage[margin=1in]{geometry}

\title{PLaMo-100B: A Ground-Up Language Model Designed for Japanese Proficiency}

\author{
 \parbox{\linewidth}{
 \centering
 \textbf{Preferred Elements, Inc.},
\footnote{
Authors listed alphabetically by last name. Please cite as Preferred Elements et al.
}
\\ 
Kenshin~Abe,
Kaizaburo~Chubachi,
Yasuhiro~Fujita,
Yuta~Hirokawa,
Kentaro~Imajo,
Toshiki~Kataoka,
Hiroyoshi~Komatsu,
Hiroaki~Mikami,
Tsuguo~Mogami,
Shogo~Murai,
Kosuke~Nakago,
Daisuke~Nishino,
Toru~Ogawa,
Daisuke~Okanohara,
Yoshihiko~Ozaki,
Shotaro~Sano,
Shuji~Suzuki,
Tianqi~Xu,
Toshihiko~Yanase
 }
}

\affil{\texttt {plamo-100b-report@preferred.jp}}

\date{}

\begin{document}

\maketitle

\begin{abstract}

We introduce PLaMo-100B, a large-scale language model designed for Japanese proficiency. 
The model was trained from scratch using 2 trillion tokens, with architecture such as QK Normalization and Z-Loss to ensure training stability during the training process. 
Post-training techniques, including Supervised Fine-Tuning and Direct Preference Optimization, were applied to refine the model's performance. 
Benchmark evaluations suggest that PLaMo-100B performs well, particularly in Japanese-specific tasks, achieving results that are competitive with frontier models like GPT-4.
The base model is available at \url{https://huggingface.co/pfnet/plamo-100b}.
\end{abstract}

\section{Introduction}

Large language models (LLMs) have revolutionized natural language processing (NLP) by enabling sophisticated solutions to complex linguistic tasks. As part of the Generative AI Accelerator Challenge (GENIAC) project\footnote{\url{https://www.meti.go.jp/english/policy/mono_info_service/geniac/index.html}} in 2024, we developed \emph{PLaMo-100B}, a 100 billion parameter LLM, which represents a major advancement in this field, particularly for its strong performance in both Japanese and English. Unlike models that are fine-tuned from pre-existing model weights like LLaMA-3~\citep{Llama-Team-AI-Meta2024techreport}, PLaMo-100B was trained from scratch, tailored specifically for optimal performance in Japanese language tasks, while also delivering high proficiency in English.

The pre-training process for PLaMo-100B used 2 trillion tokens, with 1.5 trillion for initial pre-training and 0.5 trillion for continued pre-training. The dataset was carefully curated from diverse sources, including RefinedWeb~\citep{neurips_refinedweb} and CommonCrawl\footnote{\url{https://commoncrawl.org/}}.
The model's architecture is based on a decoder-only transformer \citep{Vaswani2017NeurIPS}, incorporating advanced techniques such as QK Normalization~\citep{henry-etal-2020-query} and Z-Loss~\citep{jmlr_palm}. The training process leveraged 3D parallelism~\citep{arxiv_megatron}, Zero Bubble technique~\citep{iclr_zerobubble}, and FP8 training to efficiently handle the massive scale of the model, using NVIDIA H100 GPUs for computation.

Post-training involved a multi-stage approach to refine the model's capabilities. This included Supervised Fine-tuning (SFT) using a variety of instruction-following datasets, followed by multiple rounds of Direct Preference Optimization (DPO)~\citep{Rafael2023,xu2024cringe} to align the model's outputs with human preferences.
We also employed model merging techniques to combine different model checkpoints effectively.
A key feature of the post-training process was the extensive use of synthetic data to expand the model's capabilities and address the scarcity of high-quality datasets, particularly for the Japanese language.

PLaMo-100B has been evaluated using benchmarks such as Jaster~\citep{Han_llm-jp-eval_2024}, 
Japanese MT-Bench~\citep{zheng2023judging,japanese-mt-bench}, and the Rakuda Benchmark~\citep{rakuda-bench}.
These assessments indicate competitive performance in both Japanese and English language tasks. 
Notably, PLaMo-100B-Instruct has achieved superior performance to GPT-4-0125-Preview on Jaster and Rakuda Benchmark, demonstrating its proficiency in Japanese language tasks.

This white paper details the development of PLaMo-100B, covering its architecture, training methodology, and benchmark performance. By sharing our approach, we hope to contribute to the ongoing advancement of LLM technology and highlight progress in AI development within Japan.

\section{Pre-Training}

Our base pre-trained model, PLaMo-100B-Base, has been trained on 2 trillion tokens of text data in Japanese and English.
The training corpus primarily consists of publicly available datasets, supplemented by data that we have curated through our web crawling efforts.
In the following sections, we will delineate the training data and expound upon the model training process.

\subsection{Dataset}

For PLaMo-100B-Base, the training was conducted in two phases: the initial phase encompassed 1.5 trillion tokens, while the subsequent phase included 0.5 trillion tokens. \Cref{tab:pretraining_dataset} shows the ratio of the datasets.

The English component required 1.3 trillion tokens, whereas the Japanese component necessitated 0.7 trillion tokens for pre-training. Notably, the collection of sufficiently high-quality Japanese datasets exclusively from publicly available sources posed significant challenges. Consequently, akin to the methodologies employed in CCNet \citep{lrec_ccnet} and RefinedWeb \citep{neurips_refinedweb}, we constructed a Japanese dataset by using web-crawled data archived by CommonCrawl.

\begin{table}[t]
      \centering
      \caption{
      The ratio of datasets used for pre-training by language (including programming languages in the English category)
      }
      \label{tab:pretraining_dataset}
      \begin{tabular}{lrrrr}
\toprule
 & RefinedWeb & Other English Dataset & Proprietary CommonCrawl-JP & Other Japanese Dataset \\
\midrule
1.5T token & 42\% & 28\% & 18\% & 12\% \\
0.5T token & 17\% & 33\% & 46\% & 4\% \\
\bottomrule
\end{tabular}

\end{table}

\subsubsection{Japanese Dataset}

Initially, we constructed the Japanese corpus from CommonCrawl using CCNet. However, for the latter portion consisting of 0.5 trillion tokens, we undertook a comprehensive preprocessing effort from the ground up, using 20 data dumps spanning from 2017 to 2024, thereby generating a dataset comprising approximately 460 billion tokens.

There are two main reasons for this approach:
\begin{itemize}
    \item The WET files processed by CCNet lack the structured information that is typically available in the HTML or Markdown format; consequently, we opted to process the WARC files directly.
    \item Accumulated expertise during this project has bolstered our confidence that we can efficiently generate datasets independently.
\end{itemize}

The processing pipeline, akin to RefinedWeb \citep{neurips_refinedweb} and the Swallow Corpus\footnote{\url{https://tokyotech-llm.github.io/swallow-corpus}}, managed the raw archived data stored in the WARC format through the following steps:
\begin{enumerate}
    \item Download WARC files while extracting Japanese HTML and text files.
    \item Convert the extracted data to Markdown if it is in HTML format.
    \item Filter the entire document using llm-jp corpus filter.
    \item Conduct deduplication across all dumps employing the MinHash algorithm.
    \item Re-shard the data into uniformly sized chunks, suitable for dataset consumption.
\end{enumerate}

CommonCrawl comprises collections of segmented files ranging in size from several GiB, thereby facilitating embarrassingly parallel processing with the exception of the MinHash step. For this task, we executed each process in parallel using 1\,000 instances. Steps 1 through 3 required approximately 30 hours per dump, step 4 took about 24 hours, and step 5 was completed within several hours.

\subsection{Model Training Stabilization}

Stabilizing model training is a significant challenge in the pre-training of LLMs.
\citet{jmlr_palm} indicated that as LLMs increase in size, their training processes tend to exhibit greater instability.
In the training of PLaMo-100B-Base, we addressed this challenge using two primary strategies: the model architecture and the loss function.

The effectiveness of these strategies in our pre-training remains uncertain. Although the pre-training was executed without any observable instability, we cannot conclusively attribute this success to the aforementioned strategies; it is possible that the training setup possessed inherent stability. To disentangle these factors, it would be necessary to perform an additional pre-training session of equivalent scale, which is too costly.
Nonetheless, given the complexities associated with re-running large-scale pre-training, we contend that the implementation of proactive measures was essential.

\subsubsection{QK Normalization}

The architecture of PLaMo-100B closely resembles that of Llama2\footnote{\url{https://ai.meta.com/research/publications/llama-2-open-foundation-and-fine-tuned-chat-models/}} and Llama3\footnote{\url{https://ai.meta.com/blog/meta-llama-3/}}.
To enhance training stability, we implemented QK Normalization \citep{iclr_small_scale_proxies} because
\citet{iclr_small_scale_proxies} indicate that QK Normalization effectively stabilizes computations within self-attention layers and contributes to the overall stability of model training.

\begin{figure}
      \centering
      \includegraphics[width=6cm]{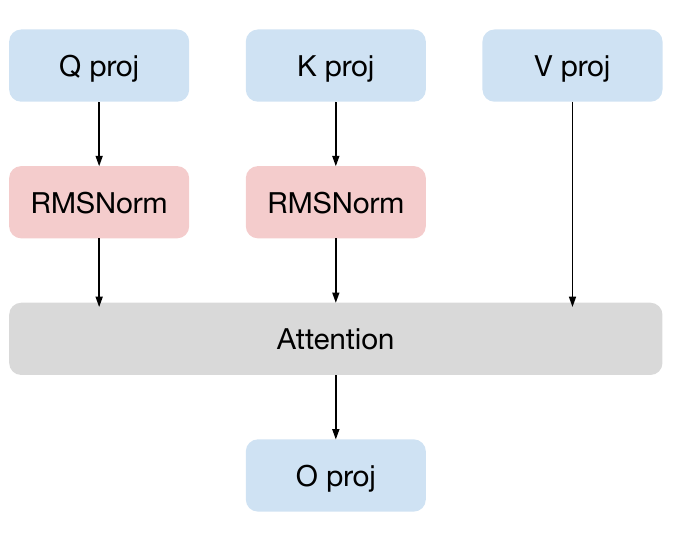}
      \caption{Self-attention layers with QK normalization}
      \label{fig:qk_normalization}
\end{figure}

In preliminary experiments, we verified that QK Normalization does not adversely affect model performance. Additionally, findings in other studies, such as Jamba \citep{arxiv_jamba} and Chameleon \citep{arxiv_chameleon}, demonstrate that incorporating a normalization layer prior to the interaction between tokens enhances training stability. This observation suggests that QK Normalization may emerge as a standard technique for large-scale models.

\subsubsection{Z-loss}

Regarding the loss function, we incorporated z-loss, which enhances numerical stability within the softmax cross-entropy loss function, defined as
\begin{equation}
    L(x) = \left(\log \left(\sum_{i=0}^{C} \exp \left(x[i]\right) \right) \right)^2,
\end{equation}
where $x \in \mathbb{R}^C$ is the output of PLaMo-100B for the next token prediction and $C$ is the vocabulary size.
Similar to QK Normalization, \citet{iclr_small_scale_proxies} demonstrated that z-loss contributes to the stabilization of training processes.

In preliminary experiments, we verified that z-loss does not adversely affect model performance, akin to QK Normalization. While it remains uncertain whether z-loss achieves its intended purpose of stabilizing training, we have yet to observe any negative impact from its implementation.
Furthermore, z-loss has proven to be a valuable metric for monitoring training progress. In instances where training deviates from expectations (such as due to a bug), the changes in z-loss are often more significant than those observed in other loss functions or in downstream task performance. This characteristic facilitates the identification of whether observed changes are attributable to trial-to-trial variability or other factors.

\subsubsection{Negative Results}

While we have implemented two methods that contributed to the stabilization of training, we also explored additional techniques. In this section, we discuss those methods that either exhibited no discernible effect or had a counterproductive impact on training stability.

\paragraph{Parallel Layers:}
In the case of our previous pre-training model, PLaMo-13B \citep{PLaMo2023Introducing}, we implemented Parallel Layers \citep{acl_workshop_parallel} as a technique to improve computational speed. Initially, there was negligible difference in performance with or without the implementation of Parallel Layers. However, as development advanced, the negative impact on model performance attributed to the use of Parallel Layers became increasingly apparent. As a result, we opted to revert to the standard Transformer architecture for PLaMo-100B-Base.

\paragraph{Normalization of Embeddings:}
\citet{iclr_8bit_optimizer} have reported that normalizing the output of embedding layers can stabilize training and enhance the performance of LLMs. We also evaluated this method, yielding mixed results:
\begin{itemize}
    \item Perplexity: A slight improvement in perplexity was observed.
    \item Generation: We noted a degradation in performance for tasks requiring decoding.
\end{itemize}
One could argue that, in the context of pre-training, perplexity should be prioritized over word generation tasks, which are primarily relevant during instruction tuning. Nonetheless, we assessed the potential risks associated with this method to be too significant and ultimately decided against its adoption in our training process.

\paragraph{Sequence Length Warmup:}
Sequence length warmup \citep{neurips_slw} may enhance training efficiency and improve model performance. This technique involves commencing training with shorter sequence lengths and gradually increasing the sequence length as the training progresses.
Upon evaluation, we determined that sequence length warmup had little effect in our experiments. This method was originally proposed to address the challenges associated with large batch sizes. However, according to \citet{arxiv_scaling_lm}, larger models are capable of using larger batch sizes without encountering significant issues. Therefore, in our context, the implementation of sequence length warmup was deemed unnecessary.

\subsection{Performance Optimization}

As of February 2024, when we began the pre-training of PLaMo-100B-Base, there were no established cases of training LLMs in the 100 billion parameter range using H100 GPUs. While employing H100 GPUs for training, minor overheads that were negligible with A100 GPUs are anticipated to become problematic. This is attributed to the H100's capacity for exceptionally high computational speeds (FLOP/s), which may cause non-computationally-bound operations to consume relatively more time.

Throughout the training phase, we implemented various strategies aimed at enhancing training speed. Some of these strategies were informed by publicly available resources, such as the efficient utilization of FP8 TensorCore. However, we also encountered several challenges that could only be discerned during the actual training process.
In this section, we present two approaches that emerged from our specific experiences.

\subsubsection{ZeRo Bubble}
For pre-training, we implemented 3D parallelism \citep{arxiv_megatron}, a method that integrates data parallelism with two types of model parallelism: tensor parallelism and pipeline parallelism to enable the training of large-scale models.

As for pipeline parallelism, we adopted Zero Bubble \citep{iclr_zerobubble}. It is recognized that pipeline parallelism may encounter inefficiencies due to periods in which certain GPUs remain idle, known as ``bubbles.'' However, Zero Bubble aims to effectively minimize these idle periods to zero.

We did not implement the speculative parameter updates introduced by Zero Bubble for the following reasons:
\begin{itemize}
    \item The definition of one iteration becomes ambiguous in the Python script, complicating the debugging process.
    \item Gradient clipping was consistently applied, resulting in very few iterations where speculative execution could be deemed effective.
\end{itemize}
The second point stands in contrast to the findings reported in the original paper, which suggested that gradient clipping is infrequently applied. We posit that this discrepancy may stem from differences in model size between our setup and that used in the study.

\subsubsection{Numerical Precision of the lm-head (Linear Layer for Word Prediction)}

LLMs, including PLaMo-100B-Base, are composed of repeated Transformer blocks but necessitate a final linear layer to predict the next token. In models provided via Hugging Face's Transformers library \citep{Wolf2020ACLDemo}, this layer is commonly referred to as the lm-head.

\begin{figure}
      \centering
      \includegraphics[width=16cm]{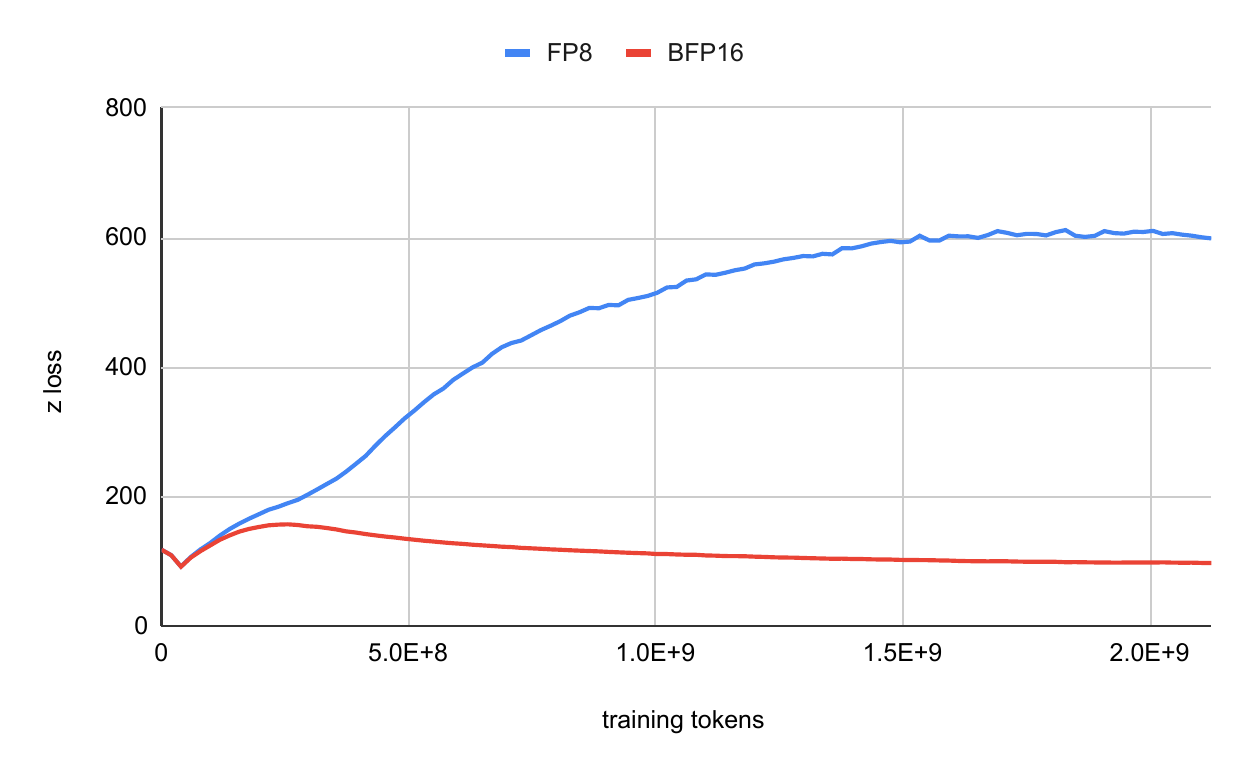}
      \caption{Difference in the z-loss values between using bfloat16 and FP8 for the lm-head.}
      \label{fig:z-loss}
\end{figure}

Initially, we computed this layer in FP8 format to enhance training speed.
Although we observed minimal issues concerning training loss, we experienced suboptimal performance in subsequent benchmark tasks.
Through investigations conducted in smaller experimental settings, we discovered that the z-loss values were significantly elevated when using FP8 for the lm-head as shown by \Cref{fig:z-loss}, suggesting that the lm-head should be computed in a higher precision format than FP8 format.
In our case, we used bfloat16 for the lm-head to mitigate this issue.

\subsubsection{Runtime Performance}
The pretraining of PLaMo-100B-Base achieved a computational speed of approximately 540 TFLOP/s/GPU, which is about 27\% of the theoretical speed of 1\,979 TFLOP/s for FP8 on the H100. While a direct comparison is difficult due to differences in the number of GPUs used, we believe that our performance is comparable to that of Llama3 and the benchmarking by MosaicML\footnote{\url{https://github.com/mosaicml/llm-foundry/tree/main/scripts/train/benchmarking}}.

\section{Post-Training}

\subsection{Post-training pipeline}

\begin{figure}
      \centering
      \includegraphics[width=\textwidth]{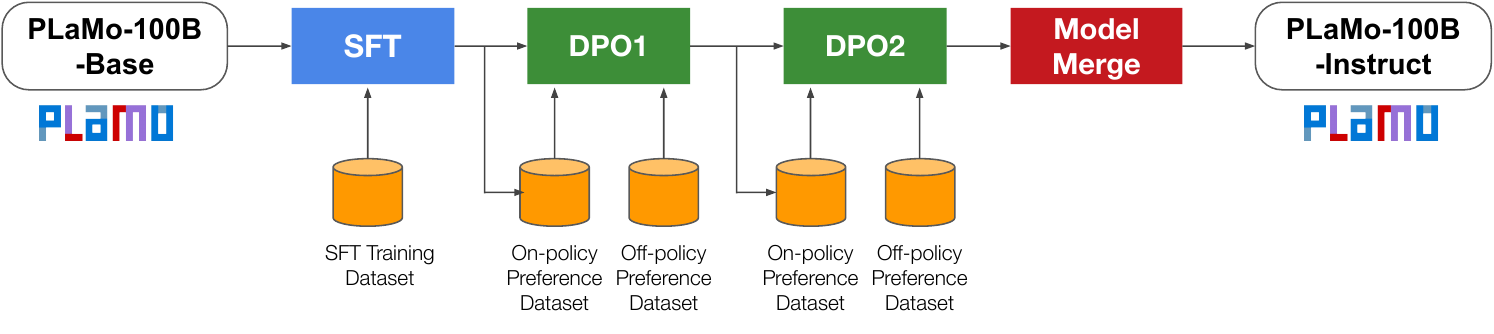}
      \caption{Summary diagram of the training pipeline.}
      \label{fig:alignment-pipeline}
\end{figure}

In our post-training phase, we employed Supervised Fine-Tuning (SFT) and Direct Preference Optimization (DPO) algorithms. 
As outlined in the subsequent sections, we implemented a training pipeline that consists of SFT followed by Iterative DPO. 
Furthermore, we adopted the Model Merging technique to enhance generalization performance post-training. 
A summary of the training pipeline is presented in \Cref{fig:alignment-pipeline}.

\subsubsection{Supervised Fine-Tuning}

In Supervised Fine-Tuning (SFT), the training process is guided by a predefined dataset comprising paired examples of ``questions posed to the LLM" and their respective ``expected answers from the LLM". 
This dataset encompasses a variety of tasks, each consisting of a question-answer pair. 
For instance, in the domain of mathematical problem-solving, each pair entails a problem and its respective solution. 
Similarly, in conversational settings, each pair consists of a user prompt and the preferred response. 
The objective of using a training dataset of high diversity and quality is to enhance the downstream task performance of the LLM.

Training in SFT is conducted using next-token prediction, analogous to the pre-training phase, albeit with an emphasis on fine-tuning the response generation component. 
To optimize this process, only the response portion of the input text is considered during loss calculation rather than the entire input sequence. 
Although our preliminary experimental evaluations revealed no significant performance difference between this approach and that considers the entire input, this method was adopted to mitigate the risk of the model internalizing undesirable traits, such as aggressive expressions, potentially present in the questions.

Traditionally, in SFT, various tasks' training datasets are aggregated and trained simultaneously. 
However, some prior studies, such as Nemotron-4 \citep{Nemotron4}, have documented conflicts arising from concurrent learning of multiple tasks. 
Despite attempts to alleviate these conflicts through adjustments to the sample weighting ratios, attaining a substantial level of harmony was challenging, especially in coding tasks.
Consequently, a two-stage SFT approach is proposed wherein the coding tasks are trained in the first stage of SFT, followed by the second stage of SFT, which focuses on more general tasks.

In our experiment, a similar trend was observed for mathematical questions. 
Therefore, we adopted the two-stage SFT method, initially segregating mathematical questions for the first stage and subsequently addressing various other tasks in the second stage.

\subsubsection{Direct Preference Optimization}

Direct Preference Optimization (DPO) is an algorithm proposed by \citet{Rafael2023} that learns human preferences from labeled data, wherein pairs of responses to the same question are designated as either better (chosen) or worse (rejected). 
The model is encouraged to generate more preferable responses by leveraging this preference information.

In the learning process of DPO, several existing datasets, such as \texttt{hh-rlhf} \citep{bai2022constitutionalai} and \texttt{HelpSteer2} \citep{wang2024helpsteer2}, which include labeled responses generated by LLMs or human authors, are commonly used. 
Alternatively, data can also be synthesized by allowing the model to generate multiple response candidates, which can then be labeled accordingly. 
The former scenario, where the model producing the responses differs from the one being trained, is referred to as off-policy, while the latter scenario, in which the model being trained generates the responses, is termed on-policy.

On-policy training has demonstrated effectiveness in enhancing learning efficiency, as it provides preference feedback on the types of responses the model is more likely to generate \citep{tajwar2024}. 
A hybrid approach, known as SPIN \citep{chen2024spin}, has also been introduced, which involves generating a dataset by pairing teacher responses from the SFT dataset with model-generated responses, under the assumption that the teacher's responses are more preferred.

When employing on-policy data for training, it is possible to alternate between data generation and DPO training termed Iterative DPO. 
This method has been shown to yield superior results compared to a single round of DPO \citep{xu2024cringe, dong2024rlhfworkflow}.

Three different datasets are combined for our two-stage DPO training after SFT: (1) a publicly available dataset, (2) a dataset generated by labeling responses produced from a snapshot of PLaMo-100B, and (3) a dataset generated through the SPIN methodology. 
This approach enables us to take advantage of both high-quality publicly available preference datasets and the efficacy of Iterative DPO using on-policy datasets. 
The details of the data generation process are described in the data generation section.

\subsubsection{Model Merging}

Model merging is a technique that integrates multiple models to enhance overall performance \citep{izmailov2019averagingweights, wortsman2022modelsoup}. 
Various methodologies exist for model merging; for instance, Llama-3.1 \citep{Llama-Team-AI-Meta2024techreport} has reported using the average of multiple models. 
We employed a straightforward model merging technique known as Spherical Linear Interpolation (SLERP) \citep{slerp}, which computes the midpoint between two models.

There were several DPO training results, depending on the combination of training data and other factors such as hyperparameters. 
By merging two of these models with distinct characteristics, we were able to create a model that incorporates the strengths of both models to some extent.

\subsection{Post-training Data}

To effectively conduct post-training, curating a diverse and high-quality dataset that describes the desired responses to user inquiries is necessary.
In the early days of post-training, InstructGPT \citep{ouyang2022instructgpt} engaged annotators to assemble a dataset that specifies the expected behavior of large language models in response to user questions.
However, as LLM development has progressed, there have been efforts to have LLMs themselves construct post-training datasets \citep{bai2022constitutionalai, wang2023selfinstruct}.
Since manually creating datasets can be costly, we decided to focus on developing data generation methods that can also be scalable in the future.

In practice, we explored the following methodologies for post-training:
\begin{enumerate}
  \item Using public data
  \item Generating data programmatically
  \item Generating data using LLMs
\end{enumerate}
We refrained from using any data that restricts commercial use, as well as outputs from proprietary models such as GPT-4 or GPT-3.5.
We also excluded the Jaster training data due to its prohibition under GENIAC guidelines.

\subsubsection{Public Data}

High-quality post-training datasets suitable for commercial use are available, such as \texttt{oasst2} \citep{kopf2023oasst} or \texttt{hh-rlhf} \citep{bai2022constitutionalai} in English, along with \texttt{ichikara-instruction}\footnote{\url{https://huggingface.co/datasets/p1atdev/ichikara-instruction}} in Japanese. 
Furthermore, the quantity of publicly accessible datasets is continuously increasing. 
We conducted experiments on a variety of these datasets to decide our instruction-tuning dataset collections.

\subsubsection{Programatic Data Generation}

To accurately address mathematical problems, we developed templates for various problem types that require calculations, subsequently generating datasets by varying the numerical values. Our mathematics dataset was manually constructed without machine learning techniques. 
Although there is a limit to the number of problem templates that can be created manually, and many data points would have only different numerical values, we decided it would be okay based on our previous empirical studies and the considerations outlined below.

When an LLM produces a calculation result, the distribution of tokens tends to exhibit a deterministic quality, which aligns with our objectives during the training process. 
Furthermore, the textual content outside of the mathematical formulas is likely to adhere to a standardized format. 
Even when the only difference between data points is the numerical values, some degree of diversity in outcomes may still be observed. 
This includes distinctions such as whether carrying occurs during addition, the potential for simplification of fractional results, or the choice of which variable to eliminate in simultaneous linear equations.

Existing datasets that do not rely on machine learning include the AMPS pretraining corpus \citep{hendrycks2021} and the work by \citet{saxton2018analysing}. 
However, these datasets feature artificial \TeX{} representations for formulas, and their answers are restricted to numerical values, indicating potential areas for enhancement in post-training applications. 
To address this, we generated our own mathematical datasets aimed at augmenting the volume of Japanese mathematical data. 
We have used the math-related datasets for pre-training as well, but for post-training, we apply a different format, such as instruction-based responses, and combine different datasets with varying ratios, taking into account the characteristics of each dataset.

\subsubsection{Synthetic Data Generation}

For the question-answering dataset, we employed the Self-Instruct algorithm \citep{wang2023selfinstruct} as a foundation for data generation. 
However, rather than using the algorithm directly with GPT, we developed a method to facilitate the data generation using smaller LLMs like PLaMo-13B \citep{PLaMo2023Introducing}.
For instance, when attempting to generate a question sentence directly, the results were suboptimal, prompting us to incorporate an additional step to first generate a concise title.

During the development of PLaMo-100B, we also focused on translating the collected and generated datasets into Japanese. 
The availability of post-training datasets for commercial use in Japanese is severely limited, creating challenges in acquiring a sufficient quantity and diversity of training data. 
Even when generating our own data, numerous open LLMs are primarily designed in English, complicating the generation of high-quality Japanese responses. 
By using PLaMo-100B itself for translation, we successfully increased the volume of high-quality Japanese data, resulting in performance enhancements in Japanese text generation tasks.

\subsubsection{Preference Data Generation}

We generated preference data during the post-training process of PLaMo-100B. 
Referring to the work by \citet{dong2024rlhfworkflow}, we generated eight different responses for the same prompt using PLaMo-100B and evaluated their scores.
The highest-scoring response was selected as the ``chosen" response, while the lowest-scoring one was marked as ``rejected". 
To evaluate response scores, we experimented with both the LLM-as-a-Judge method using open LLMs \citep{zheng2023judging,verga2024poll} and the reward model.

In this data generation process, only the prompt is required, and a teacher response example is unnecessary.
We can can datasets like \texttt{chatbot\_arena\_conversations} \citep{zheng2023judging}, which only contains user prompts in a commercially usable license.
During the response generation where LLM inference is required, we used vLLM \citep{kwon2023efficient} for acceleration.

\section{Evaluation Results}

For evaluating the model, we used the \texttt{g-leaderboard} branch of the \texttt{llm-leaderboard} benchmark\footnote{\url{https://github.com/wandb/llm-leaderboard/tree/g-leaderboard}} operated by Weights \& Biases on the GENIAC 1.0, and measured Jaster and MT-Bench.
Additionally, we also used in-house evaluation code to measure the Rakuda Benchmark.

\subsection{Jaster}

Jaster is a collection of Japanese benchmarks, to measure the ability of LLMs to understand Japanese. 
It is evaluated using the code from the \texttt{llm-jp-eval} repository.
In the GENIAC project, it is evaluated on the \texttt{g-leaderboard} branch\footnote{\url{https://github.com/llm-jp/llm-jp-eval/tree/g-leaderboard}} using a specific set of categories such as
\begin{enumerate}
  \item NLI: Natural Language Inference
  \item QA: Question Answering
  \item RC: Reading Comprehension
  \item MC: Multi-Choice QA
  \item MR: Math Reasoning
  \item FA: Fundamental Analysis.
\end{enumerate}
Each benchmark's score instead of the average score of each category is shown in \Cref{sec:jaster}.

The performance of LLMs is assessed in both 4-shot and 0-shot settings with the question-answering task.
In the 4-shot setting, examples of questions and answers are provided when asking questions, while in the 0-shot setting, no examples are given.

\begin{table}[t]
      \centering
      \caption{
          Evaluation Results for Jaster 0-shot. Note that PLaMo-100B-Instruct evaluations were conducted in-house, while other models' evaluations were based on the results provided by Weights \& Biases.
      }
      \label{tab:jaster0shot}
      \begin{tabular}{lrrrrrrr}
\toprule
Model Name & AVG & FA & MC & MR & NLI & QA & RC \\
\midrule
GPT-4 (0125 Preview) & 0.722 & 0.255 & $\mathbf{0.960}$ & $\mathbf{0.970}$ & 0.772 & $\mathbf{0.569}$ & 0.808 \\
GPT 3.5 Turbo & 0.567 & 0.183 & 0.610 & 0.770 & 0.590 & 0.429 & 0.818 \\
Swallow-70b-instruct-hf & 0.576 & 0.175 & 0.590 & 0.710 & 0.642 & 0.480 & 0.856 \\
PLaMo-100B-Base & 0.542 & 0.185 & 0.830 & 0.280 & 0.682 & 0.426 & 0.846 \\
PLaMo-100B-Instruct & $\mathbf{0.738}$ & $\mathbf{0.579}$ & 0.950 & 0.780 & $\mathbf{0.838}$ & 0.394 & $\mathbf{0.886}$ \\
\bottomrule
\end{tabular}

\end{table}

\begin{table}[]
      \centering
      \caption{
          Evaluation Results for Jaster 4-shot. Note that PLaMo-100B-Instruct evaluations were conducted in-house, while other models' evaluations were based on the results provided by Weights \& Biases.
      }
      \label{tab:jaster4shot}
      \begin{tabular}{lrrrrrrr}
\toprule
Model Name & AVG & FA & MC & MR & NLI & QA & RC \\
\midrule
GPT-4 (0125 Preview) & 0.772 & 0.405 & 0.950 & $\mathbf{0.980}$ & 0.806 & 0.623 & 0.871 \\
GPT 3.5 Turbo & 0.656 & 0.355 & 0.900 & 0.840 & 0.544 & 0.423 & 0.877 \\
Swallow-70b-instruct-hf & 0.676 & 0.365 & 0.900 & 0.770 & 0.506 & $\mathbf{0.634}$ & 0.878 \\
PLaMo-100B-Base & 0.679 & 0.305 & 0.930 & 0.610 & 0.710 & $\mathbf{0.634}$ & $\mathbf{0.884}$ \\
PLaMo-100B-Instruct & $\mathbf{0.775}$ & $\mathbf{0.592}$ & $\mathbf{0.960}$ & 0.800 & $\mathbf{0.856}$ & 0.561 & 0.881 \\
\bottomrule
\end{tabular}

\end{table}

As shown in \Cref{tab:jaster0shot} and \Cref{tab:jaster4shot}, the PLaMo-100B-Instruct model, which is obtained after post-training, significantly improved its performance compared to the base model, surpassing GPT-4's average score.
Although the Jaster training dataset was not used in this experiment, the model learned how to use the knowledge gained during pre-training by learning how to answer various question formats during post-training \citep{longpre2023flan}.
This capability contributes to the model's improved performance.
The results confirm that the PLaMo-100B model, which is trained on a higher fraction of Japanese data, has a strong foundation in Japanese language understanding.

The only category where the model's performance fell short of GPT-4 was the Mathematical Reasoning (MR) category.

We constructed a large-scale math dataset as explained in the previous section and performed extensive SFT.
However, achieving near-perfect accuracy in this category requires a high level of generalization in mathematical problem-solving abilities, which might not be fully achieved during pre-training without sufficient exposure to a wide variety of mathematical examples.

\subsection{MT-Bench}

MT-Bench \citep{zheng2023judging} serves as a benchmark for assessing the conversational response capabilities of LLMs, evaluating the quality of responses across eight categories: coding, extraction, humanities, math, reasoning, roleplay, stem and writing.
Since the responses are free-form conversations, rule-based scoring methods are impractical.
Instead, the LLM-as-Judge approach is employed, using models such as GPT-4 as evaluators to assign scores.
While the original MT-Bench is presented in English, Stability AI has developed a Japanese version, which is publicly accessible as Japanese MT-Bench \citep{japanese-mt-bench}.

Within the GENIAC project, we have evaluated the scores of both the English and Japanese versions of MT-Bench and reported in \Cref{tab:mtbenchja,tab:mtbenchen}.

\begin{table}[t]
      \centering
      \caption{
          Evaluation Results for Japanese MT-Bench. Note that PLaMo-100B-Instruct evaluations were conducted in-house, while other models' evaluations were based on the results provided by Weights \& Biases.
      }
      \label{tab:mtbenchja}
      \begin{tabular}
{@{\hspace{6pt}}l@{\hspace{6pt}}r@{\hspace{6pt}}r@{\hspace{6pt}}r@{\hspace{6pt}}r@{\hspace{6pt}}r@{\hspace{6pt}}r@{\hspace{6pt}}r@{\hspace{6pt}}r@{\hspace{6pt}}r@{\hspace{6pt}}r@{\hspace{6pt}}}
\toprule
Model Name & AVG & coding & extraction & humanities & math & reasoning & roleplay & stem & writing \\
\midrule
GPT-4 (0125 Preview) & $\mathbf{8.93}$ & $\mathbf{9.10}$ & 8.50 & 8.55 & $\mathbf{8.65}$ & $\mathbf{8.25}$ & 9.00 & $\mathbf{9.80}$ & $\mathbf{9.55}$ \\
GPT 3.5 Turbo & 8.00 & 8.40 & $\mathbf{8.65}$ & $\mathbf{9.75}$ & 5.15 & 6.35 & 8.75 & 8.50 & 8.45 \\
Mixtral-8x7B-Instruct-v0.1 & 6.98 & 6.65 & 8.15 & 8.50 & 5.00 & 6.05 & 6.40 & 8.40 & 6.65 \\
Swallow-70b-instruct-hf & 4.38 & 3.35 & 5.70 & 4.80 & 1.45 & 3.75 & 5.85 & 5.85 & 4.30 \\
PLaMo-100B-Base & 5.47 & 3.85 & 4.65 & 7.60 & 3.95 & 2.55 & 8.00 & 7.70 & 5.45 \\
PLaMo-100B-Instruct & 7.78 & 4.95 & 7.70 & $\mathbf{9.75}$ & 4.90 & 7.35 & $\mathbf{9.25}$ & 9.10 & 9.25 \\
\bottomrule
\end{tabular}

\end{table}

\begin{table}[]
      \centering
      \caption{
          Evaluation Results for MT-Bench (English). Note that PLaMo-100B-Instruct evaluations were conducted in-house, while other models' evaluations were based on the results provided by Weights \& Biases.
      }
      \label{tab:mtbenchen}
      \begin{tabular}{@{\hspace{6pt}}l@{\hspace{6pt}}r@{\hspace{6pt}}r@{\hspace{6pt}}r@{\hspace{6pt}}r@{\hspace{6pt}}r@{\hspace{6pt}}r@{\hspace{6pt}}r@{\hspace{6pt}}r@{\hspace{6pt}}r@{\hspace{6pt}}r@{\hspace{6pt}}}
\toprule
Model Name & AVG & coding & extraction & humanities & math & reasoning & roleplay & stem & writing \\
\midrule
GPT-4 (0125 Preview) & $\mathbf{9.02}$ & $\mathbf{8.05}$ & $\mathbf{8.90}$ & $\mathbf{10.00}$ & $\mathbf{7.45}$ & $\mathbf{8.80}$ & $\mathbf{9.40}$ & $\mathbf{9.90}$ & 9.65 \\
GPT 3.5 Turbo & 8.40 & 7.10 & 8.70 & 9.75 & 6.95 & 7.75 & 8.75 & 9.12 & 9.10 \\
Mixtral-8x7B-Instruct-v0.1 & 8.02 & 6.15 & 8.30 & 9.90 & 4.60 & 7.30 & 8.97 & 9.15 & $\mathbf{9.80}$ \\
Swallow-70b-instruct-hf & 4.87 & 3.05 & 7.00 & 7.20 & 1.45 & 3.70 & 5.05 & 6.10 & 5.37 \\
PLaMo-100B-Base & 5.76 & 3.35 & 3.82 & 8.85 & 4.20 & 4.55 & 7.70 & 7.67 & 5.95 \\
PLaMo-100B-Instruct & 7.70 & 4.20 & 7.60 & 9.90 & 5.75 & 6.35 & 9.25 & 9.45 & 9.10 \\
\bottomrule
\end{tabular}

\end{table}

Based on the evaluation, PLaMo-100B-Instruct managed to score an average of 7.781 in Japanese MT-Bench, coming close to GPT-3.5 in terms of score.
Compared to the baseline model's score of 5.469, there is a significant improvement, demonstrating that the post-training process successfully enhanced the conversational response capabilities as intended.

Analyzing by category, we found that the model particularly excelled in the humanities, STEM, writing, and roleplay categories, with notable performance in responses requiring creative and open-ended dialogue.
\Cref{tab:mtbench_humanities_example} in \Cref{sec:mtbench} illustrates an example response from the humanities category.

On the other hand, the pre-trained model demonstrated subpar performance in categories requiring scientific consideration such as math, coding, and reasoning. 
As indicated by the Jaster benchmark, enhancing the model's proficiency in these domains is likely essential during the pre-training phase. 
However, we investigated how much the model's scientific capabilities could be enhanced through post-training.

As previously noted, we developed a scalable data generation method to create a substantial number of high-quality mathematics and coding datasets for post-training. 
Following training on these datasets, the model's performance in mathematics, coding, and reasoning categories exhibited significant improvement, although it still fell short of GPT-4's performance. 
However, after the post-training process, the model's abilities were found to be comparable to those of GPT-3.5 Turbo in these areas.

It is important to mention that during the post-training, user prompts from the \texttt{chatbot\_arena\_conversations} dataset were used.
We identified that eight English MT-Bench prompts were inadvertently included in this dataset.
Reference answers and Japanese MT-Bench prompts were not part of the dataset. 
Since the leak was discovered after the conclusion of post-training, the dataset containing these prompts was not removed.

\subsection{Rakuda Benchmark}

The Rakuda Benchmark is designed to assess the performance of conversational responses to questions pertaining to Japanese domestic topics, including geography, politics, history, and society.
In this study, we employed judge prompts from MT-Bench to conduct an absolute evaluation, scoring the responses on a scale of 10.
Additionally, we performed a relative evaluation using prompts provided by the Rakuda Benchmark's official guidelines.
However, we did not compute ratings based on pairwise comparisons of multiple models, which were done in official evaluations.

\begin{table}[t]
      \centering
      \caption{
          Absolute Evaluation Results for Rakuda Benchmark.
      }
      \label{tab:rakudabenchabs}
      \begin{tabular}{lrrrrr}
\toprule
Model Name & AVG & Geography & Politics & History & Society \\
\midrule
PLaMo-100B-Instruct & $\mathbf{9.73}$ & $\mathbf{9.70}$ & 9.60 & $\mathbf{9.80}$ & $\mathbf{9.80}$ \\
GPT-4-0125-Preview & 9.55 & 9.60 & $\mathbf{9.75}$ & 9.50 & 9.35 \\
GPT-4-0613 & 9.38 & 9.40 & 9.60 & 9.50 & 9.00 \\
GPT-3.5-Turbo-0301 & 8.88 & 8.50 & 9.10 & 9.00 & 8.90 \\
\bottomrule
\end{tabular}

\end{table}

\begin{table}[t]
      \centering
      \caption{
          Pairwise Evaluation Results for Rakuda Benchmark.
          It is known that the order of presentation may bias the pairwise evaluation, so we evaluated 40 questions * 2 orders of presentation.
      }
      \label{tab:rakudabenchpair}
      \begin{tabular}{ll}
\toprule
Model pair & Win-loss records \\
\midrule
PLaMo-100B-Instruct vs. GPT-4-0125-Preview & 42 wins, 36 losses, 2 ties \\
PLaMo-100B-Instruct vs. GPT-4-0613 & 58 wins, 21 losses, 1 tie \\
\bottomrule
\end{tabular}

\end{table}

The maximum score for the absolute evaluation is 10 points.
Notably, the results generated by PLaMo-100B-Instruct did not contain any errors detectable by the judge model.
The primary variations in scores were attributed to the evaluation of the answer details.
In the domain of geography, knowledge proved to be particularly significant, and it appears that the volume of Japanese data used during pre-training contributed to the wide coverage of this knowledge.
Conversely, in the other three domains, the judge model frequently emphasized critical aspects such as ``perspective", ``impact" and ``challenges", which resulted in challenges in achieving high relative evaluation scores in comparison to GPT-4-0125-Preview.

Here are some notes regarding the evaluation process.
Response generation for PLaMo and the absolute evaluation were conducted using in-house implementations to facilitate faster inference, without altering the generation parameters.
The judge model was specified as GPT-4, and the evaluation used the GPT-4-0613 version from Azure OpenAI, which was the latest version available at the time of the final update of the Rakuda Benchmark's official evaluation.
The response data for GPT-4 0125-Preview was generated using the code provided in the Rakuda Benchmark repository.
In contrast, the responses for GPT-4-0613 and GPT-3.5-Turbo-0301 were obtained from the published response data available in the Rakuda Benchmark repository, with the GPT-4 version estimated based on the date (20230713) indicated in the file names.

\section{Conclusion}

With the computational resources provided by GENIAC, we successfully completed the pre-training and post-training of PLaMo-100B, a large-scale language model. PLaMo-100B-Instruct achieved notable performance, surpassing GPT-4 in Japanese-specific benchmarks such as Jaster and Rakuda, though areas like mathematical reasoning and coding still need improvement.

The project also led to the creation of a scalable pipeline for dataset generation and training, which can be applied to other models beyond PLaMo. This provides a strong foundation for future model development and broader applications.

During GENIAC project, other teams in our company explored multi-modality extensions using the PLaMo-100B to enjoy its Japanese proficiency.
Concretely, an image adaptor was trained to support various visual tasks, e.g., object detection, image classification, caption generation, and visual question answering.
As another direction, \citet{nozawa2024plamo100b_speech} fine-tuned PLaMo-100B for Japanese automatic speech recognition with a pre-trained audio encoder and an additional adapter.

Furthermore, we are actively considering the safety and ethical implications of large language models. For more details on our stance on responsible technology development, please refer to our statement ``Responsibility/責任ある技術開発に向けて"\footnote{\url{https://www.preferred.jp/ja/company/responsibility/}}.

\section*{Author Contributions}

Within each section, contributors are listed in alphabetical order by last name.

\paragraph*{Pre-training team}

Yuta~Hirokawa,
Hiroyoshi~Komatsu,
Hiroaki~Mikami,
Shogo~Murai,
Daisuke~Nishino,
Shuji~Suzuki,
Tianqi~Xu

\paragraph*{Post-training team}

Kenshin~Abe,
Kaizaburo~Chubachi,
Yasuhiro~Fujita,
Kentaro~Imajo,
Toshiki~Kataoka,
Tsuguo~Mogami,
Kosuke~Nakago,
Toru~Ogawa,
Yoshihiko~Ozaki,
Toshihiko~Yanase

\paragraph*{Overall project management}

Daisuke~Okanohara,
Shotaro~Sano

\section*{Acknowledgments}

This paper is based on results obtained from the project, ``Research and Development Project of the Enhanced Infrastructures for Post 5G Information and Communication System'' (JPNP 20017), subsidized by the New Energy and Industrial Technology Development Organization (NEDO).

We would like to thank PFN/PFE members, especially Daisuke Tanaka, Kento Nozawa, Naoki Imazato, and Toru Taniguchi for their helpful discussions and/or implementation support.
We also would like to thank cluster team members such as Akira Kanai, Hidehito Yabuuchi, Hirochika Asai, Hiroki Okui, Hirotaka Akita, Kazuki Suda, Keiko Okada, Kengo Suga, Kiyoshi Mizumaru, Kohei Sugihara, Moto Ishizawa, Sho Shimizu, Takeshi Senoo, Tetsuya Shiota, Toru Komatsu, Yojiro Uo, Yuichiro Ueno, and Yusuke Doi for the infrastructure support.

\bibliography{reference}
\bibliographystyle{plainnat}

\newpage
\appendix

\section{Jaster Detailed Evaluation Results}
\label{sec:jaster}

Jaster evaluation results for each benchmark are shown in \Cref{tab:mtbenchja,tab:mtbenchen} for 0-shot and 4-shot respectively.

\begin{table}[t]
      \centering
      \caption{
          Detailed Evaluation Results for Jaster 0-shot. Note that PLaMo-100B-Instruct evaluations were conducted in-house, while other models' evaluations were based on the results provided by Weights \& Biases.
      }
      \label{tab:jaster0shot_detail}
      \small
\begin{tabular}{
@{\hspace{6pt}}l
@{\hspace{6pt}}r
@{\hspace{6pt}}r
@{\hspace{6pt}}r
@{\hspace{6pt}}r
@{\hspace{6pt}}r
@{\hspace{6pt}}r
@{\hspace{6pt}}r
@{\hspace{6pt}}r
@{\hspace{6pt}}r
@{\hspace{6pt}}r
@{\hspace{6pt}}r
@{\hspace{6pt}}r
@{\hspace{6pt}}r
@{\hspace{6pt}}r
@{\hspace{6pt}}r}
\toprule
Model Name & \begin{turn}{90}jamp\end{turn} & \begin{turn}{90}janli\end{turn} & \begin{turn}{90}jcommonsenseqa\end{turn} & \begin{turn}{90}jemhopqa\end{turn} & \begin{turn}{90}jnli\end{turn} & \begin{turn}{90}jsem\end{turn} & \begin{turn}{90}jsick\end{turn} & \begin{turn}{90}jsquad\end{turn} & \begin{turn}{90}mawps\end{turn} & \begin{turn}{90}niilc\end{turn} & \begin{turn}{90}wiki\_coreference\end{turn} & \begin{turn}{90}wiki\_dependency\end{turn} & \begin{turn}{90}wiki\_ner\end{turn} & \begin{turn}{90}wiki\_pas\end{turn} & \begin{turn}{90}wiki\_reading\end{turn} \\
\midrule
GPT-4 (0125 Preview) & 0.59 & 0.98 & 0.96 & 0.53 & 0.78 & 0.76 & 0.75 & 0.81 & 0.97 & 0.61 & 0.02 & 0.13 & 0.15 & 0.02 & 0.95 \\
GPT 3.5 Turbo & 0.47 & 0.69 & 0.61 & 0.45 & 0.49 & 0.61 & 0.69 & 0.82 & 0.77 & 0.41 & 0.03 & 0.01 & 0.00 & 0.00 & 0.87 \\
Swallow-70b-instruct-hf & 0.48 & 0.85 & 0.59 & 0.36 & 0.55 & 0.69 & 0.64 & 0.86 & 0.71 & 0.60 & 0.00 & 0.00 & 0.00 & 0.00 & 0.88 \\
PLaMo-100B-Base & 0.48 & 0.72 & 0.83 & 0.26 & 0.77 & 0.73 & 0.71 & 0.85 & 0.28 & 0.59 & 0.03 & 0.00 & 0.00 & 0.00 & 0.89 \\
PLaMo-100B-Instruct & 0.90 & 1.00 & 0.95 & 0.25 & 0.78 & 0.73 & 0.78 & 0.89 & 0.78 & 0.54 & 0.38 & 0.94 & 0.10 & 0.52 & 0.95 \\
\bottomrule
\end{tabular}
\end{table}

\begin{table}[t]
      \centering
      \caption{
          Detailed Evaluation Results for Jaster 4-shot. Note that PLaMo-100B-Instruct evaluations were conducted in-house, while other models' evaluations were based on the results provided by Weights \& Biases.
      }
      \label{tab:jaster4shot_detail}
      \small
\begin{tabular}{
@{\hspace{6pt}}l
@{\hspace{6pt}}r
@{\hspace{6pt}}r
@{\hspace{6pt}}r
@{\hspace{6pt}}r
@{\hspace{6pt}}r
@{\hspace{6pt}}r
@{\hspace{6pt}}r
@{\hspace{6pt}}r
@{\hspace{6pt}}r
@{\hspace{6pt}}r
@{\hspace{6pt}}r
@{\hspace{6pt}}r
@{\hspace{6pt}}r
@{\hspace{6pt}}r
@{\hspace{6pt}}r}
\toprule
Model Name & \begin{turn}{90}jamp\end{turn} & \begin{turn}{90}janli\end{turn} & \begin{turn}{90}jcommonsenseqa\end{turn} & \begin{turn}{90}jemhopqa\end{turn} & \begin{turn}{90}jnli\end{turn} & \begin{turn}{90}jsem\end{turn} & \begin{turn}{90}jsick\end{turn} & \begin{turn}{90}jsquad\end{turn} & \begin{turn}{90}mawps\end{turn} & \begin{turn}{90}niilc\end{turn} & \begin{turn}{90}wiki\_coreference\end{turn} & \begin{turn}{90}wiki\_dependency\end{turn} & \begin{turn}{90}wiki\_ner\end{turn} & \begin{turn}{90}wiki\_pas\end{turn} & \begin{turn}{90}wiki\_reading\end{turn} \\
\midrule
GPT-4 (0125 Preview) & 0.70 & 1.00 & 0.95 & 0.64 & 0.77 & 0.80 & 0.76 & 0.87 & 0.98 & 0.60 & 0.07 & 0.62 & 0.20 & 0.19 & 0.95 \\
GPT 3.5 Turbo & 0.48 & 0.80 & 0.90 & 0.42 & 0.43 & 0.48 & 0.53 & 0.88 & 0.84 & 0.42 & 0.03 & 0.52 & 0.15 & 0.16 & 0.92 \\
Swallow-70b-instruct-hf & 0.32 & 0.76 & 0.90 & 0.64 & 0.57 & 0.62 & 0.26 & 0.88 & 0.77 & 0.63 & 0.12 & 0.47 & 0.20 & 0.09 & 0.94 \\
PLaMo-100B-Base & 0.41 & 0.84 & 0.93 & 0.60 & 0.85 & 0.76 & 0.69 & 0.88 & 0.61 & 0.66 & 0.04 & 0.37 & 0.20 & 0.06 & 0.85 \\
PLaMo-100B-Instruct & 0.93 & 1.00 & 0.96 & 0.50 & 0.79 & 0.76 & 0.80 & 0.88 & 0.80 & 0.63 & 0.37 & 0.92 & 0.10 & 0.60 & 0.97 \\

\bottomrule
\end{tabular}
\end{table}

Please refer to the Nejumi LLM leaderboard Neo\footnote{\url{https://wandb.ai/wandb-japan/llm-leaderboard/reports/Nejumi-LLM-Neo--Vmlldzo2MTkyMTU0\#llm-jp-eval}} for the explanation as well as the sample questions of each benchmark.

\section{MT-Bench Response Example}
\label{sec:mtbench}

\begin{table}[t]
      \centering
      \caption{
          Example of a question and a response from the humanities category in Japanese MT-Bench, along with the corresponding score for PLaMo-100B-Instruct.
      }
      \label{tab:mtbench_humanities_example}
      \begin{tabularx}{\textwidth}{lXXr}
\hline
Turn & User & PLaMo-100B-Instruct & Score \\
\endfirsthead %
\hline
1 &
\tiny{日本でビジネスを行う際のビジネスマナーについて説明してください。} &
\tiny{日本でビジネスを行う際には、日本独自のビジネスマナーが存在します。以下に、その一部をご紹介します。

・名刺交換

日本では、名刺交換は非常に重要なビジネスマナーの一つです。名刺交換の際には、名刺を両手で受け取り、相手の目を見て挨拶を交わします。また、名刺を受け取る際には「頂戴いたします」と言い、相手の名刺を丁寧に扱います。

・挨拶

日本では、挨拶を非常に重視します。ビジネスの場では、「おはようございます」「こんにちは」「こんばんは」などの挨拶を、相手に対して丁寧に行うことが求められます。また、挨拶の際にはお辞儀をすることが一般的です。お辞儀の角度やタイミングにもマナーがありますので、注意が必要です。

・時間厳守

日本では、時間を守ることが非常に重要視されます。ビジネスの場では、約束の時間に遅れないようにすることが求められます。やむを得ず遅刻をする場合には、事前に連絡を入れることがマナーです。

・言葉遣い

日本では、敬語を使うことが一般的です。相手に対して失礼のないように、適切な敬語を使うことが求められます。また、ビジネスの場では、ネガティブな言葉や否定的な表現を避けることが望ましいとされています。

・服装

日本では、ビジネスの場での服装にもマナーがあります。男性の場合はスーツが一般的で、女性の場合はスーツやジャケット、ブラウスなどのフォーマルな服装が求められます。また、ビジネスの場では、清潔感のある服装を心がけることが大切です。

以上のように、日本でビジネスを行う際には、日本独自のビジネスマナーを理解し、適切に対応することが求められます。これらのマナーを守ることで、相手に対して敬意を表し、ビジネスを円滑に進めることができます。} &
10 \\

2 &
\tiny{日本のカーディテーリング事業の新入社員向けトレーニングのビデオスクリプトを作成してください。上記のエチケットの規範を強調してください。} &
\tiny{[オープニングシーン: 清潔でプロフェッショナルな外観のカーディテイリングショップ。トレーナーがカメラに向かって微笑みながら挨拶をする。]

トレーナー: 「こんにちは、そしてようこそ私たちのカーディテイリングビジネスへ！私は本日のトレーナーです。今日から数日間、皆さんに基本的なビジネスマナーやエチケットについての研修を行います。これらのスキルは、お客様や同僚との良好な関係を築くために非常に重要です。では、さっそく始めましょう。」

[シーン1: 名刺交換]

トレーナー: 「日本では、名刺交換は非常に重要なビジネスマナーです。名刺を交換する際には、必ず立って、相手の目を見て、両手で名刺を受け取り、渡します。また、名刺を受け取る際には、『名刺を頂戴いたします』と言い、相手の名刺を丁寧に扱います。」

[シーン2: 挨拶]

トレーナー: 「挨拶も非常に重要です。お客様や同僚に対して、『おはようございます』、『こんにちは』、『こんばんは』などの挨拶を、適切なタイミングで行うようにしましょう。また、挨拶をする際には、お辞儀をすることが一般的です。お辞儀の角度やタイミングにも気を配りましょう。」

[シーン3: 時間厳守]

トレーナー: 「時間を守ることも非常に重要です。お客様との約束や、チームミーティングなどには、必ず時間通りに、あるいは少し早めに到着するようにしましょう。もし、遅刻をしてしまう場合には、事前に連絡を入れるようにしましょう。」

[シーン4: 言葉遣い]

トレーナー: 「言葉遣いは、相手に対する敬意を表すために非常に重要です。お客様や同僚に対しては、敬語を使うようにしましょう。また、ネガティブな言葉や否定的な表現は避けるようにしましょう。」

[シーン5: 服装]

トレーナー: 「服装も、相手に対する敬意を表すために重要です。男性の場合は、スーツやジャケットを着用することが一般的です。女性の場合は、スーツやジャケット、ブラウスなどのフォーマルな服装が求められます。また、清潔感のある服装を心がけるようにしましょう。」

[クロージングシーン: 要点の振り返りと励まし]

トレーナー: 「以上が、今日の研修の主なポイントです。これらのマナーやエチケットを守ることで、お客様や同僚との良好な関係を築くことができます。最初は難しいかもしれませんが、練習を重ねることで、自然と身についていきます。皆さんの活躍を期待しています！」

[エンドシーン: 会社のロゴと連絡先情報が表示される。]} &
10 \\
\hline
\end{tabularx}

\end{table}

\end{document}